\documentclass{article}
\usepackage[preprint]{colm2026_conference}

\usepackage{microtype}
\usepackage{hyperref}
\usepackage{url}
\usepackage{booktabs}
\usepackage{amsmath}
\usepackage{amssymb}
\usepackage{graphicx}
\usepackage{tikz}
\usetikzlibrary{arrows.meta, positioning, fit, backgrounds}
\usepackage{lineno}

\definecolor{darkblue}{rgb}{0, 0, 0.5}
\hypersetup{colorlinks=true, citecolor=darkblue, linkcolor=darkblue, urlcolor=darkblue}

\title{Supersede: Diagnosing and Training the\\ Memory-Update Gap in LLM Agents}

\author{Vedant Patel \\
Vrin \,$\cdot$\, \texttt{vedant@vrin.cloud}}

\begin{document}

\ifcolmsubmission
\linenumbers
\fi

\maketitle

\begin{center}
\small\href{https://github.com/Vrin-cloud/supersede}{Code \& Environment} \,$\cdot$\, \href{https://app.primeintellect.ai/dashboard/environments/vedant/supersede}{Prime Intellect Hub} \,$\cdot$\, \href{https://huggingface.co/vedant33/supersede-qwen2.5-3b-grpo-lora}{Model \& Data}
\end{center}

\begin{abstract}
  Large language model (LLM) agents increasingly operate over long,
  multi-session interactions in which facts change: a user moves, a price
  updates, a plan is revised. Acting correctly requires using the
  \emph{current} value of a fact and discarding values that have been
  \emph{superseded}. We isolate this ability on real conversational data and
  show that it is a distinct, unsolved failure. On the
  \texttt{knowledge-update} subset of LongMemEval, replacing an agent's full
  context with a bounded, self-maintained memory drops accuracy from
  $92\%$ to $77\%$ even on a frontier model (gpt-5.4), a gap that is
  statistically significant (paired McNemar $p<0.005$) and \emph{persists
  across model scale} while full-context accuracy saturates near $92\%$. The
  bottleneck is therefore memory maintenance, not comprehension, and it is not
  closed by a stronger model. We then ask whether this is merely an artifact of
  an undersized memory, and find it is not: as the conversation grows
  $24\times$, accuracy falls further (from $68\%$ to $28\%$), and granting the
  agent proportionally more memory yields no detectable recovery
  ($28\%\!\rightarrow\!28\%$, $n=25$). The failure scales with the length of the
  conversation, not with the compression ratio. We release
  \textbf{Supersede},
  an open reinforcement-learning environment (built on the \texttt{verifiers}
  /\,\texttt{prime-rl} stack) that turns this measurement into a training
  signal: agents are rewarded for answering from the current value and
  penalized for stale ones. Finally, we close the loop and show the gap is
  trainable: GRPO fine-tuning a small open model (Qwen2.5-3B) on this
  environment nearly doubles its held-out supersession accuracy on real,
  unseen conversations ($9.0\%\!\rightarrow\!16.7\%$, a single run), along a monotonic
  checkpoint curve that indicates the learned policy, not the harness,
  carries the gain. To our knowledge this is the first trainable environment
  whose reward targets temporal fact-currency, and the first evidence that the
  supersession gap can be trained down rather than only measured.
\end{abstract}

\section{Introduction}

Pre-training has largely consumed the open web, and the frontier of LLM
capability has shifted toward agents that act over long horizons using
\emph{external memory} rather than a single context window. A defining
property of long-horizon interaction is that information is not static: facts
are introduced, then later updated or retracted. An assistant that is told a
user lives in Detroit, and later that they moved to the suburbs, must answer a
subsequent location question with the \emph{current} value. We call the
correct handling of such updates \emph{supersession}, and the accuracy an agent
loses when it must sustain it under a bounded, self-maintained memory the
\emph{supersession gap}. This paper measures that gap, shows the obvious
remedies do not close it, and then trains it down.

Recent benchmarks have begun to measure long-term memory
\citep{maharana2024locomo, wu2025longmemeval, memoryarena2026,
memoryagentbench2025, fama2026}, and a forgetting-aware metric has been
proposed to penalize reliance on obsolete information \citep{fama2026}. These
works, however, score \emph{frozen} models: they quantify the problem but do
not provide a mechanism to \emph{train} it away. In parallel, a line of work
applies reinforcement learning to memory agents \citep{memagent2025,
longrlvr2026}, but rewards final-answer correctness or evidence relevance,
never temporal currency. No existing work sits in the intersection: a
\emph{trainable} environment whose reward is supersession-correctness.

This paper makes five contributions, structured as one investigation: we ask
in turn whether the failure exists, whether a bigger model fixes it, whether a
bigger memory fixes it, and whether \emph{training} fixes it --- and answer no,
no, no, and yes. The first three answers establish that the easy escapes are
closed; the fourth shows the principled one is open.

\begin{enumerate}
  \item \textbf{An isolation result.} On real conversational data, we hold
  memory load fixed and vary only whether the agent has full context or a
  bounded, self-maintained memory. Bounded memory degrades
  \texttt{knowledge-update} accuracy significantly (Section~\ref{sec:gap}),
  isolating memory maintenance, not reading comprehension, as the cause.
  \item \textbf{A frontier confirmation.} The gap survives on gpt-5.4
  ($92\%\!\rightarrow\!77\%$, $p=0.0033$) and does not shrink to zero as the
  model improves; full-context accuracy, by contrast, saturates. A bigger model
  does not fix it.
  \item \textbf{Scale, not size.} As the conversation grows $24\times$, the
  failure deepens ($68\%\!\rightarrow\!28\%$), and giving the agent
  proportionally more memory yields no detectable recovery at $n=25$
  (Section~\ref{sec:scale}). The failure tracks the length of the conversation,
  not the compression ratio, so a bigger memory does not fix it either.
  \item \textbf{An open environment.} We release \textbf{Supersede}\footnote{Code, environment, trained model, and dataset are publicly
  available: \url{https://github.com/Vrin-cloud/supersede} (environment),
  \url{https://huggingface.co/vedant33/supersede-qwen2.5-3b-grpo-lora} (model), and
  \url{https://huggingface.co/datasets/vedant33/supersede-rl-episodes} (dataset); the
  environment is also published on the Prime Intellect Environments Hub.}, a
  \texttt{verifiers}/\texttt{prime-rl} environment with a programmatic
  supersession-aware reward, validated end-to-end. It reframes the
  forgetting-aware \emph{metric} of \citet{fama2026} as a \emph{learning
  signal}, so the failure can be trained against rather than only measured.
  \item \textbf{Training closes it.} Fine-tuning a small open model
  (Qwen2.5-3B) on the environment with GRPO nearly doubles held-out
  supersession accuracy on real, unseen conversations
  ($9.0\%\!\rightarrow\!16.7\%$, Section~\ref{sec:train}; a single training run, multi-seed significance in ongoing work), monotonically as
  training progresses. Where a bigger model and a bigger memory did not, a
  trained memory policy does, turning the diagnosis into a fix.
\end{enumerate}

\section{Related Work}

\paragraph{Long-term memory benchmarks.} LoCoMo \citep{maharana2024locomo}
evaluates very long multi-session dialogue but does not annotate a dedicated
update category. LongMemEval \citep{wu2025longmemeval} introduces an explicit
\texttt{knowledge-update} question type in which a fact stated in one session
is changed in a later one; we adopt this subset as ground truth. MemoryArena
\citep{memoryarena2026} shows that models near-perfect on recall benchmarks
drop sharply when memory must drive actions, motivating evaluation beyond
passive recall. MemoryAgentBench \citep{memoryagentbench2025} and MemBench
\citep{membench2025} add knowledge-updating and conflict-resolution competencies. Memora and its FAMA metric \citep{fama2026}
explicitly penalize reliance on superseded or deleted memory, and temporal-conflict
QA \citep{temporalconflict2026} probes whether models resolve facts that change over
time. All of these are evaluation-only.

\paragraph{Memory systems and RL for memory.} Mem0 \citep{chhikara2025mem0}
manages a memory store with prompted \textsc{add}/\textsc{update}/\textsc{delete}
operations: a heuristic, not a learned, policy. A line of work instead
\emph{learns} memory management with RL: MemAgent \citep{memagent2025} rewards
final-answer correctness, LongRLVR \citep{longrlvr2026} rewards evidence
\emph{relevance} within a single long document, AgeMem \citep{agemem2026} learns
store/update/discard operations under \emph{task-outcome} rewards, and Memory-T1
\citep{memoryt12025} learns to \emph{select} temporally relevant sessions from the
full history. Two things separate Supersede. First, these methods may re-read the
raw history; Supersede forbids re-feeding, so the only state is the agent's own
bounded memory. Second, their reward is task success, evidence relevance, or
temporal consistency, not whether the agent kept the \emph{current} value of a
changed fact. Supersede makes that the reward itself: a verifiable, time-indexed
supersession signal that reframes FAMA's forgetting-aware \emph{objective} \citep{fama2026} as a
dense training target. The closest contemporary method, MemPO \citep{mempo2026},
also pairs GRPO with a self-managed bounded memory, but rewards memory
task-sufficiency (credit assignment over memory effectiveness), not which value
is current. Put differently, prior RL-for-memory work learns
\emph{what to answer}; Supersede learns \emph{which version is current}: the
supersession signal is the reward, not a proxy for it.

\paragraph{Deployed memory systems.} Memory reached production across every major
lab in 2026: OpenAI's ``Dreaming'' \citep{openai2026dreaming}, Anthropic's Claude
memory \citep{anthropic2026memory}, and Google's Gemini personalization
\citep{gemini2026memory} all run background processes that consolidate and rewrite
a user's memory to keep facts current. Tellingly, OpenAI's own time-sensitive
evaluation of updating outdated context self-reports only $75.1\%$ success (up
from $9.4\%$ in 2024), and Gemini's documentation notes that profile updates can
lag by days. These systems are closed, with no released benchmark, methodology,
or trainable signal: the gap Supersede fills.

\paragraph{Environment infrastructure.} We build on the \texttt{verifiers}
library and the Prime Intellect Environments Hub
\citep{primeintellect2025environments}, which standardize environments for
post-training, and are compatible with the emerging OpenEnv standard
\citep{openenv2025}.

\section{The Supersede Environment}
\label{sec:env}

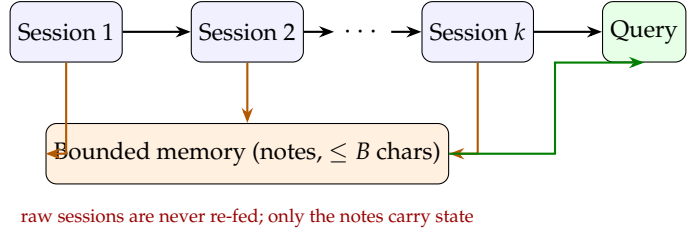
\begin{figure}[t]
  \centering
  \begin{tikzpicture}[
    node distance=6mm and 9mm,
    box/.style={draw, rounded corners, align=center, font=\small,
                minimum height=8mm, inner sep=3pt},
    sess/.style={box, fill=blue!6},
    mem/.style={box, fill=orange!12},
    q/.style={box, fill=green!10},
    arr/.style={-{Stealth[length=2mm]}, thick},
  ]
    \node[sess] (s1) {Session 1};
    \node[sess, right=of s1] (s2) {Session 2};
    \node[right=4mm of s2] (dots) {$\cdots$};
    \node[sess, right=4mm of dots] (sk) {Session $k$};
    \node[q, right=of sk] (qq) {Query};

    \node[mem, below=8mm of s2] (m) {Bounded memory (notes, $\le B$ chars)};

    \draw[arr] (s1) -- (s2);
    \draw[arr] (s2) -- (dots);
    \draw[arr] (dots) -- (sk);
    \draw[arr] (sk) -- (qq);

    \draw[arr, orange!70!black] (s1) |- (m.west);
    \draw[arr, orange!70!black] (s2) -- (m);
    \draw[arr, orange!70!black] (sk) |- (m.east);
    \draw[arr, green!50!black] (m.east) -- ++(1.4,0) |- (qq.south);

    \node[font=\scriptsize, text=red!60!black, below=2mm of m]
      {raw sessions are never re-fed; only the notes carry state};
  \end{tikzpicture}
  \caption{The Supersede rollout. The agent rewrites a bounded notes memory
  after each session and answers the final query from memory alone. Because
  history is not re-fed, a superseded value that is not overwritten persists
  and corrupts the answer.}
  \label{fig:rollout}
\end{figure}

\paragraph{Task.} A task is a multi-session interaction followed by a query
about the current value of a fact that changed during the interaction. The
agent processes one session at a time and maintains a \emph{bounded} memory
(a notes field capped at $B$ characters); crucially, raw sessions are
\emph{never re-fed}. After the final session, the agent answers the query using
its memory alone. Figure~\ref{fig:rollout} shows the rollout.

\paragraph{Reward.} The primary reward is $r_{\text{cur}}=\mathbb{1}[\text{the
final answer conveys the current value}]$, scored by a programmatic,
ungameable matcher (normalized variant match with a token-overlap fallback),
so the environment can be evaluated and trained without an auxiliary judge
model. When the superseded values are known (our synthetic tasks, or
update-annotated items), a penalty
$r_{\text{stale}}=-\mathbb{1}[\text{the answer asserts a superseded value}]$
is added; on gold-only data it is inert. The combined reward is
$r = r_{\text{cur}} + \lambda\, r_{\text{stale}}$. The runs reported here optimize
$r_{\text{cur}}$ alone (weight $1.0$); the stale penalty enters only in the
environment's ablation. This makes the
forgetting-aware accuracy of \citet{fama2026} a dense training signal rather
than a post-hoc score.

\paragraph{Data.} Tasks are drawn from the LongMemEval
\texttt{knowledge-update} subset (real conversational supersession) and from a
procedural generator of synthetic timelines used for controls and ablations.
The environment is implemented as a \texttt{verifiers} \texttt{MultiTurnEnv};
the prompt assembly enforces the no-re-feed property, and rollouts terminate
when the answer is produced.

\section{Experimental Setup}

\paragraph{Why real, not synthetic, data.} We first built a procedural
generator of templated supersession timelines (explicit ``$X$ now $Y$''
updates). With the history in context, frontier models score $100\%$ across
configurations: they resolve clean, explicit updates by a last-mention scan.
Synthetic templated supersession is therefore \emph{saturated} and cannot
surface the failure, which is likely why it is under-measured. We accordingly
use real conversational data throughout, where updates are implicit and
paraphrased. The generator is retained as a control and for the
environment's stale-penalty ablation. As Section~\ref{sec:train} shows, it also
serves as a training curriculum whose learned skill transfers back to the real data,
turning a saturated probe into a useful teacher.

\paragraph{Data and grading.} We use the LongMemEval \texttt{knowledge-update}
subset. Section~\ref{sec:gap} uses all $78$ questions on the \emph{oracle} split
(evidence sessions only, $\sim$2 sessions/question), graded by an LLM judge
(\texttt{gpt-4.1-mini}) following the LongMemEval protocol. Because this judge is
itself one of the evaluated models, we cross-check it against the judge-free
programmatic matcher (Section~\ref{sec:scale}), which agrees to within a few
points, guarding against self-grading bias. Section~\ref{sec:scale} uses the same
questions on the much longer \emph{$\_s$} split ($\sim$48 sessions, $\sim$122k
tokens/question), graded by the programmatic matcher held constant across
conditions for a fair comparison. The environment's intrinsic reward uses the
same matcher.

\paragraph{Conditions.} \textsc{Full-context} places every session in the model
context and asks the question (an upper bound where memory is not the
bottleneck). \textsc{Bounded-memory} is the Supersede agent of
Section~\ref{sec:env}: a notes field of $B$ characters ($B{=}300$ unless noted),
rewritten one session at
a time, with raw sessions never re-fed. For the scale study we vary both the
history length (oracle vs.\ $\_s$) and the budget $B$, including a
\emph{constant-ratio} setting in which $B$ grows proportionally with the number
of sessions ($150$ characters per session, matching the oracle ratio).

\paragraph{Models and significance.} gpt-4.1-mini and gpt-4.1 at temperature $0$, and the frontier reasoning model
gpt-5.4 at \texttt{reasoning\_effort}{=}low (no temperature). Conditions are run
on the same questions, so we report a paired McNemar test (continuity-corrected
$\chi^2$).

\paragraph{Training setup.} The diagnosis above uses proprietary frontier
models, which we cannot fine-tune; to test whether the environment's reward is
a usable \emph{learning} signal we therefore switch to an open, trainable
model, Qwen2.5-3B-Instruct \citep{qwen2025}. We train with GRPO
\citep{shao2024deepseekmath}, which estimates its baseline from a group of
sampled rollouts rather than a learned value critic (a natural fit for our
programmatic, verifiable reward), on the \texttt{prime-rl} stack (LoRA, rank
$32$, $\alpha\,{=}\,64$ \citep{hu2022lora}; learning rate $10^{-5}$; batch $32$,
group $8$; $4\times$ A10G GPUs). Training tasks come from the procedural generator in
\emph{bounded-memory} mode (a fixed set of $2000$ episodes, disjoint from the real
held-out questions) --- $6$--$8$ session timelines in which an
introduced fact is later superseded and the update is buried among distractor
sessions. Two points make this a genuine test rather than a tautology. First,
the held-out evaluation is the \emph{real} LongMemEval oracle set
(Section~\ref{sec:gap}), never seen in training, so we measure transfer from
synthetic to real supersession, not memorization. Second, the synthetic
saturation noted above is a property of \emph{frontier} models reading the
\emph{full} history by last-mention scan; for a small model that must instead
\emph{maintain} the fact in a bounded memory, even templated supersession is an
unsolved behavior to learn (its untrained reward is far from ceiling,
Section~\ref{sec:train}). Base and trained models are graded by the same
programmatic matcher used in Section~\ref{sec:scale}, so the pre/post comparison
is internally exact. All held-out evaluations (base and every checkpoint) use
greedy decoding (temperature~$0$), so a checkpoint's score is deterministic under
re-evaluation: re-running the eval reproduces the same count, and the only
stochastic factor in the held-out result is the single training seed, which we
flag in Section~\ref{sec:limits}.

\section{Results}

\subsection{Bounded memory degrades supersession}
\label{sec:gap}

\begin{figure}[t]
  \centering
  \includegraphics[width=0.85\linewidth]{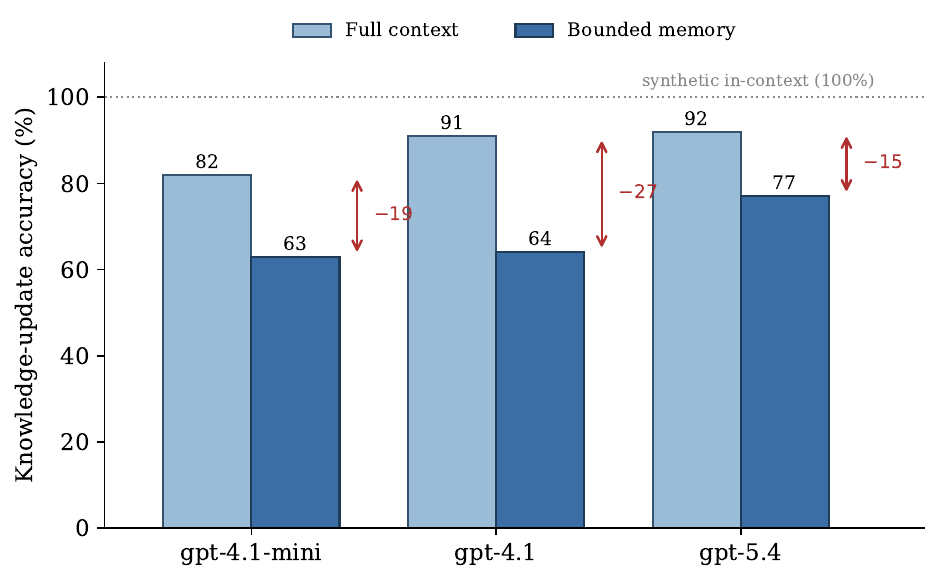}
  \caption{Knowledge-update accuracy ($n=78$ oracle questions, LongMemEval
  judge protocol): full context vs.\ bounded memory, across
  three models. Stronger models read context better (full-context rises to
  $92\%$) but bounded-memory accuracy does not catch up; this supersession gap
  is the cost of memory maintenance under updates. The dotted line marks
  synthetic in-context
  performance ($100\%$), which is saturated and uninformative.}
  \label{fig:results}
\end{figure}

\begin{table}[t]
  \caption{LongMemEval \texttt{knowledge-update}, oracle split ($n=78$). The
  full-context\,$\rightarrow$\,bounded-memory gap is significant on both the
  small and the frontier model. (McNemar is not computed for the intermediate
  gpt-4.1; the small and frontier models bracket the effect.)}
  \label{tab:main}
  \centering
  \begin{tabular}{lccc}
    \toprule
    Model & Full context & Bounded memory & Paired McNemar \\
    \midrule
    gpt-4.1-mini & $82\%$ & $63\%$ & $p=0.0035$ \\
    gpt-4.1      & $91\%$ & $64\%$ & --- \\
    gpt-5.4      & $92\%$ & $77\%$ & $p=0.0033$ \\
    \bottomrule
  \end{tabular}
\end{table}

\paragraph{The gap is significant, and a bigger model does not close it.}
Table~\ref{tab:main} and Figure~\ref{fig:results} report the result. For
gpt-4.1-mini, accuracy falls from $82\%$ to $63\%$: the paired test counts $19$
questions full-context answers correctly but bounded-memory does not, against
only $4$ in reverse ($p=0.0035$). On the frontier, gpt-5.4 raises full-context
accuracy to $92\%$ but bounded-memory only to $77\%$ ($p=0.0033$; $13$ vs.\
$1$). Across the family, bounded-memory accuracy ($63, 64, 77$) climbs far more
slowly than full-context accuracy ($82, 91, 92$), which saturates. The
bottleneck is memory maintenance, not comprehension, and scaling the model only
partially relieves it. (Roughly $13\%$ of questions are wrong even in full
context, confirming real updates are genuinely hard.)

\paragraph{Failure modes.} Errors are dominated by the relevant fact being
compressed away or not overwritten. Asked ``How many Korean restaurants have I
tried?'' (gold: \emph{four}), the agent answers ``you haven't mentioned any.''
Asked ``Where did Rachel move to?'' (gold: \emph{the suburbs}), gpt-5.4 answers
``no information about Rachel.'' The model is competent; its memory no longer
contains the updated fact.

\subsection{It is scale, not memory size}
\label{sec:scale}

\begin{figure}[t]
  \centering
  \includegraphics[width=0.85\linewidth]{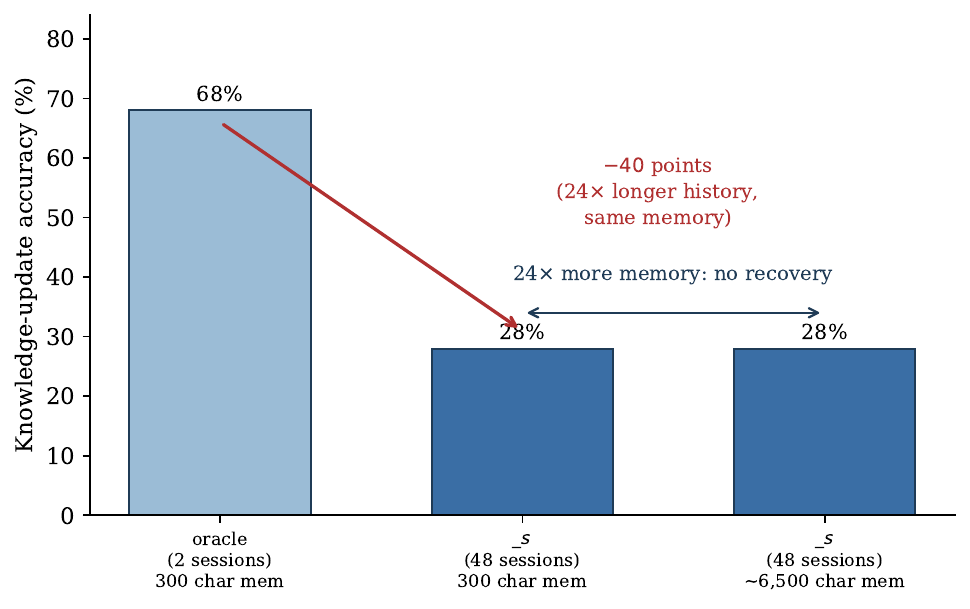}
  \caption{Scale, not size (gpt-4.1-mini, $n=25$, programmatic matcher held
  constant across conditions). Growing the conversation $24\times$ at a fixed
  $300$-character memory collapses accuracy ($68\%\!\rightarrow\!28\%$). Giving
  the agent $24\times$ more memory (constant ratio) recovers none of it
  ($28\%\!\rightarrow\!28\%$). The failure tracks conversation length, not the
  compression ratio.}
  \label{fig:scale}
\end{figure}

A natural objection to Section~\ref{sec:gap} is that $B=300$ characters is
simply too small, that the failure is an artifact of an undersized memory
rather than of supersession. We test this directly by growing the conversation
$24\times$ (oracle\,$\rightarrow\!\_s$) and, separately, growing the memory
proportionally. Results (gpt-4.1-mini, $n=25$) are in Table~\ref{tab:scale} and
Figure~\ref{fig:scale}.

\begin{table}[t]
  \caption{Scale study, LongMemEval \texttt{knowledge-update} ($n=25$,
  gpt-4.1-mini). More history collapses accuracy; proportional memory does not
  restore it.}
  \label{tab:scale}
  \centering
  \begin{tabular}{lccc}
    \toprule
    Split & History & Memory & Accuracy \\
    \midrule
    oracle & $\sim$2 sessions & $300$ char & $68\%$ \\
    $\_s$  & $\sim$48 sessions & $300$ char & $28\%$ \\
    $\_s$  & $\sim$48 sessions & $\sim$7{,}150 char (const.\ ratio) & $28\%$ \\
    \bottomrule
  \end{tabular}
\end{table}

\paragraph{More history collapses supersession.} At a fixed $300$-character
memory, lengthening the conversation from $\sim$2 to $\sim$48 sessions drops
accuracy from $68\%$ to $28\%$ (paired McNemar $p=0.002$), a $40$-point fall driven by relevant facts
being squeezed out or overwritten by stale ones across the longer history.

\paragraph{Proportional memory yields no detectable recovery ($n=25$).} Granting
the agent $24\times$ more memory (the constant-ratio setting, $\sim$7{,}150
characters) leaves accuracy unchanged at $28\%$. This is not because the budget went unused: every
one of the $25$ answers differs between the two $\_s$ conditions, so the larger
memory demonstrably changed the agent's behavior; it simply helped and hurt
in equal measure (McNemar $b=c=4$, no net effect). The failure therefore tracks
the \emph{length} of the conversation, not the compression ratio. Combined with
Section~\ref{sec:gap}, neither a bigger model nor a bigger memory closes the
gap.

\paragraph{Environment validation.} Running the released environment
end-to-end on all $78$ questions, every rollout terminates cleanly and the
intrinsic programmatic reward yields $57.7\%$ for gpt-4.1-mini, consistent with
the $63\%$ measured by the LLM judge (the matcher is stricter). The
environment thus faithfully reproduces the failure it is designed to train
against.

\subsection{Training closes the gap}
\label{sec:train}

Sections~\ref{sec:gap} and \ref{sec:scale} closed off the two easy escapes:
neither a stronger model nor a larger memory removes the gap. The remaining
hypothesis is that supersession is a \emph{policy} that must be learned. We test
it directly by training on the environment and measuring transfer to the real
held-out benchmark.

\begin{figure}[t]
  \centering
  \includegraphics[width=0.85\linewidth]{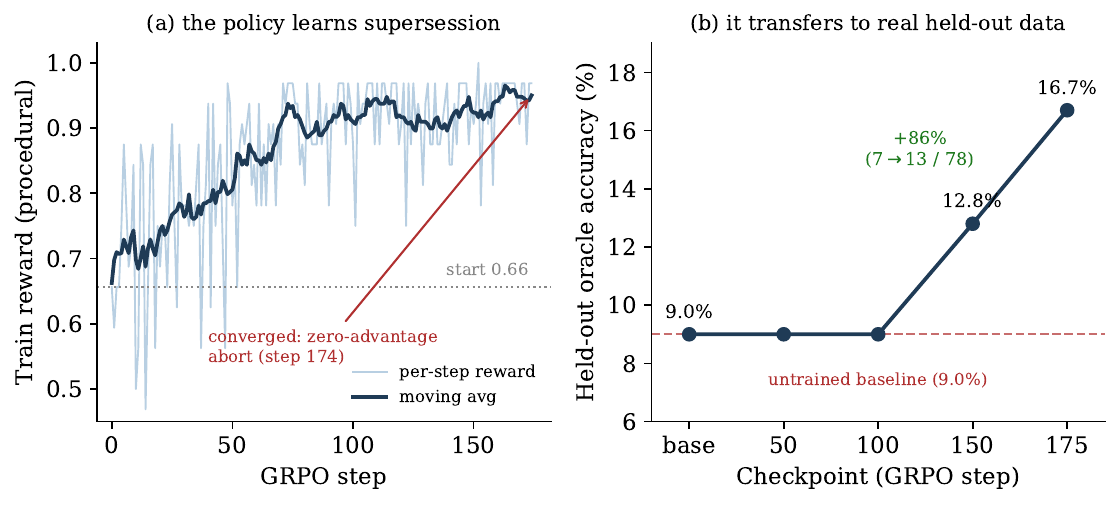}
  \caption{Training closes the gap. \textbf{(a)} GRPO reward on procedural
  episodes rises from $0.66$ to $0.97$ and then self-terminates by step $175$:
  once nearly every rollout in a group succeeds, the group-relative advantage
  vanishes and there is no gradient left: the procedural distribution is
  solved. \textbf{(b)} On the \emph{real}, held-out LongMemEval oracle set, the
  trained policy lifts accuracy from the $9.0\%$ untrained baseline to $16.7\%$.
  The rise is monotonic and switches on precisely when the reward saturates
  (\textasciitilde step $100$): the learned behavior transfers, and is still
  climbing at the final checkpoint.}
  \label{fig:train}
\end{figure}

\begin{table}[t]
  \caption{Held-out gap-closing. Qwen2.5-3B on the LongMemEval
  \texttt{knowledge-update} oracle set ($n=78$, bounded memory, programmatic
  matcher), by GRPO checkpoint. Pre and post differ only by the trained LoRA
  adapter.}
  \label{tab:train}
  \centering
  \begin{tabular}{lcc}
    \toprule
    Checkpoint & Oracle accuracy & vs.\ baseline \\
    \midrule
    base (untrained) & $7/78 = 9.0\%$  & --- \\
    step $50$        & $7/78 = 9.0\%$  & $+0.0$ \\
    step $100$       & $7/78 = 9.0\%$  & $+0.0$ \\
    step $150$       & $10/78 = 12.8\%$ & $+3.8$ \\
    \textbf{step $175$} & $\mathbf{13/78 = 16.7\%}$ & $\mathbf{+7.7}$ \\
    \bottomrule
  \end{tabular}
\end{table}

\paragraph{The policy learns the behavior.} On the procedural training
distribution, GRPO reward climbs from $0.66$ to $0.97$
(Figure~\ref{fig:train}a). The run halts itself at step $175$ on ten
consecutive zero-advantage batches: once almost every sampled rollout in a
group answers correctly, GRPO's group-relative advantage (and hence the
gradient) goes to zero. The model has exhausted what the easy episodes can
teach. This is a property of the curriculum, not a failure of training, and it
upper-bounds what this run can reach (Section~\ref{sec:limits}).

\paragraph{And it transfers to real data.} The question is whether anything
learned on synthetic timelines survives on real conversations. It does
(Table~\ref{tab:train}, Figure~\ref{fig:train}b): held-out oracle accuracy rises
from $9.0\%$ to $16.7\%$ ($13/78$ vs.\ $7/78$ correct), a $+7.7$-point ($+86\%$
relative) gain that nearly doubles the untrained baseline (a single run; see
Section~\ref{sec:limits}). The \emph{shape} of the curve is the evidence
that this is real learning and not a lucky checkpoint: accuracy is flat through
step $100$, then rises monotonically ($12.8\%$, $16.7\%$). It begins to
rise exactly when the training reward saturates, i.e.\ once the behavior is
actually acquired. Because the harness, prompt, data split, and matcher are all
held fixed, the only thing that changed is the policy; the gain is attributable
to training alone. It is still climbing at the last checkpoint before the run
self-terminated, so $16.7\%$ is a floor on what this curriculum yields, not a
ceiling.

\section{Discussion}

The results reframe supersession from a comprehension problem to a
memory-policy problem. Stronger models read updated context well, but when they
must \emph{decide what to keep} under a bounded memory, they discard or fail to
overwrite the value that later matters. This is precisely the regime that
production memory systems operate in, and precisely where a heuristic
\textsc{update}/\textsc{delete} policy \citep{chhikara2025mem0} is brittle. Our
two scaling axes rule out the easy fixes: the gap does not close as the model
grows (Section~\ref{sec:gap}), and it does not close as the memory grows
(Section~\ref{sec:scale}). Waiting for a larger model or simply allocating a
larger memory will not solve it, and allocating memory proportional to an
ever-growing conversation is itself untenable at scale. What is left is a memory
policy \emph{trained} to preserve currency, and Section~\ref{sec:train} shows
this is the axis that moves: the same bounded-memory agent, after GRPO training
on Supersede, nearly doubles its held-out accuracy. The lift is partial and the
model is small, but the contrast is the point: the two dimensions that
\emph{should} have helped (model size, memory size) do not, while the one the
environment was built to exercise (a trained currency-preserving policy) does.
Supersession is thus best understood not as a capability the next model will
absorb, but as a behavior that must be optimized for; the environment and its
verifiable reward are what make that optimization possible.

\section{Limitations}
\label{sec:limits}

The training result is a proof of mechanism, not a finished policy. It uses a
single small model (Qwen2.5-3B) and a single run; the absolute gain ($+7.7$
points) is modest and the trained accuracy ($16.7\%$) remains far below the
frontier full-context ceiling, as expected for a $3$B model and bounded
further by the curriculum. Because the procedural episodes are solved before the
run ends (training self-terminates on zero advantage, Section~\ref{sec:train}),
this curriculum cannot teach harder supersession than it contains; the clear
next lever is longer, more implicit training episodes that do not saturate, and
we expect them to extend the curve in Figure~\ref{fig:train}b rather than bend
it. What the result does establish is directional and clean: training on the
environment moves held-out accuracy monotonically, where model size and memory
size did not.

The diagnosis carries its own caveats. The scale study uses $n=25$ (vs.\ $n=78$
for the main gap), so the $40$-point drop is robust but the constant-ratio null
is measured on a smaller, noisier sample. The additional $\_s$ sessions are
distractors, so our ``scale'' axis grows the surrounding conversation rather
than the number of updates to the tracked fact; both worsen memory maintenance,
but they are not identical. Grading by LLM judge (Section~\ref{sec:gap})
introduces some noise (a few failures are judge strictness, e.g.\ ``over 25''
vs.\ gold ``25''); the paired significance survives a handful of such flips, and
the scale and training studies avoid this by using the programmatic matcher
throughout. The two diagnosis scales are themselves graded differently (the
frontier gap of Section~\ref{sec:gap} by judge, the trainable result by matcher),
so their absolute numbers are not directly comparable; what the claim rests
on is the matcher held fixed across the base/trained comparison. That matcher is
lenient on surface form (a manual audit of the trained model's matched answers
found most carry the gold value verbatim, with a few crediting near-misses);
because it grades base and trained identically, the pre/post \emph{delta} is
unaffected, though absolute accuracies should be read as matcher-graded. Finally,
the training gain is a single run: we offer the monotonic checkpoint curve as
evidence that it reflects learning rather than a lucky seed, but its statistical
significance is not formally established at $n=78$, which we address with
multi-seed runs in ongoing work.

\section{Conclusion}

Agents fail at supersession not because they cannot read updates but because
they cannot maintain them under a bounded memory. This supersession gap is
significant, reproducible on a trusted benchmark, deepens as the conversation
grows, and is closed by neither a larger model nor a larger memory. It is, however, closed in
part by \emph{training}: a small open model fine-tuned on Supersede nearly
doubles its held-out supersession accuracy, monotonically as it learns. We
release Supersede (the diagnosis, the open environment, and this first
gap-closing result together) as a step toward agents whose memory stays
current not by accident of scale, but because they were trained to keep it so.

\bibliography{refs}
\bibliographystyle{colm2026_conference}

\appendix

\section{Reward matcher}
\label{app:matcher}

The environment scores answers with a programmatic, judge-free matcher, so it can be
trained and evaluated without an auxiliary model. Given a model answer $a$ and a gold
(current) value $g$:
\begin{enumerate}
  \item \textbf{Normalize.} Both strings are lowercased, punctuation is replaced by
  spaces, and runs of whitespace are collapsed.
  \item \textbf{Gold variants.} $g$ is split on connectives (``or'', ``and'', ``/'',
  ``;'', ``,'') and each normalized piece of length $\ge 2$ is kept as an acceptable
  variant (so ``25 minutes and 50 seconds'' also accepts ``25 minutes'' or ``50 seconds'').
  \item \textbf{Match.} The answer matches if any gold variant occurs as a normalized
  substring of $a$.
  \item \textbf{Fallback.} If no variant matches, a token-overlap fallback fires: the
  answer matches when its token overlap with a gold variant exceeds $0.6$.
\end{enumerate}
$r_{\text{cur}}=\mathbb{1}[\text{match}]$. When the superseded values are known, the same
matcher is run against each stale value; a positive match there triggers the penalty
$r_{\text{stale}}$. The matcher is intentionally lenient on surface form but strict on
\emph{which} value is reported: the quantity supersession turns on.

\section{Training details}
\label{app:training}

Table~\ref{tab:hparams} lists the full GRPO configuration. Training uses procedurally
generated bounded-memory episodes ($6$--$8$ sessions, one introduced fact later superseded,
the update buried among distractor sessions); evaluation is the untouched real LongMemEval
oracle set. The run self-terminates at step $174$ after ten consecutive zero-advantage
batches: every sampled rollout in the group succeeds, so GRPO's group-relative advantage
(and the gradient) vanishes.

\begin{table}[h]
  \centering
  \caption{GRPO training configuration.}
  \label{tab:hparams}
  \begin{tabular}{ll}
    \toprule
    Setting & Value \\
    \midrule
    Base model & Qwen2.5-3B-Instruct \\
    Algorithm & GRPO \citep{shao2024deepseekmath} \\
    Adapter & LoRA \citep{hu2022lora}, rank $32$, $\alpha=64$, dropout $0$ \\
    Target modules & \texttt{\{q,k,v,o,gate,up,down\}\_proj} \\
    Bias & none \\
    Learning rate & $10^{-5}$ \\
    Batch size / group size & $32$ / $8$ \\
    Hardware & $4\times$ A10G \\
    Stack & \texttt{prime-rl} / \texttt{verifiers} \\
    Training episodes & procedural, $6$--$8$ sessions ($2000$ total) \\
    KL penalty & none ($\beta=0$) \\
    Decoding (eval) & greedy (temperature $0$); answer $\le 64$ tokens \\
    Termination & self-halt at step $175$ of $200$ max (10 zero-adv.\ batches) \\
    \bottomrule
  \end{tabular}
\end{table}

\section{Example episodes}
\label{app:examples}

Two procedurally generated training episodes. The agent sees one session at a time and
maintains a bounded notes memory; raw sessions are never re-fed. The reward is $1$ only if
the final answer conveys the \emph{current} value (in bold), not a superseded one.

\paragraph{Example 1 (single update, 7 sessions).}
\begin{itemize}\setlength\itemsep{1pt}
  \item S1: ``I recently settled in Chicago.''
  \item S2--S3: distractors (trip planning; a new book)
  \item S4: ``I moved again; I'm in Atlanta now.''
  \item S5--S7: distractors (desk; weather; a recipe)
  \item \textbf{Query:} ``Which city do I currently live in?'' --- current: \textbf{Atlanta};
  superseded: Chicago.
\end{itemize}

\paragraph{Example 2 (two updates, 6 sessions).}
\begin{itemize}\setlength\itemsep{1pt}
  \item S1: ``I got myself a Mazda.''
  \item S2--S3: distractors (trip planning; a new hobby)
  \item S4: ``New ride update: it's a Kia now.''
  \item S5: ``I switched to a Lexus now.''
  \item S6: distractor (a new book)
  \item \textbf{Query:} ``What car do I currently drive?'' --- current: \textbf{Lexus};
  superseded: Mazda, Kia.
\end{itemize}

\end{document}